\documentclass[conference]{IEEEtran}
\IEEEoverridecommandlockouts
\usepackage{cite}
\usepackage{amsmath,amssymb,amsfonts}
\usepackage{algorithmic}
\usepackage{graphicx}
\usepackage{textcomp}
\usepackage{xcolor}
\usepackage{adjustbox}
\def\BibTeX{{\rm B\kern-.05em{\sc i\kern-.025em b}\kern-.08em
    T\kern-.1667em\lower.7ex\hbox{E}\kern-.125emX}}
\begin{document}

\title{A Domain Generalization Approach for Out-Of-Distribution 12-lead ECG Classification with
Convolutional Neural Networks\\
\thanks{The work leading to these results has received funding from the European Union’s Horizon 2020 research
and innovation programme under Grant Agreement No. 965231, project REBECCA (REsearch on BrEast Cancer induced chronic
conditions supported by Causal Analysis of multi-source data).}
}

\author{\IEEEauthorblockN{Aristotelis Ballas}
\IEEEauthorblockA{\textit{Dept. of Informatics and Telematics} \\
\textit{Harokopio University of Athens}\\
Athens, Greece \\
aballas@hua.gr}
\and
\IEEEauthorblockN{Christos Diou}
\IEEEauthorblockA{\textit{Dept. of Informatics and Telematics} \\
\textit{Harokopio University of Athens}\\
Athens, Greece \\
cdiou@hua.gr}
}

\maketitle

\begin{abstract}
Deep Learning systems have achieved great success in the past few years, even surpassing human 
intelligence in several cases.
As of late, they have also established themselves in the biomedical and healthcare domains, where they have shown
a lot of promise, but have not yet achieved widespread adoption.
This is in part due to the fact that most methods fail to maintain their performance when they are called to make
decisions on data that originate from a different distribution than the one they were trained on,
namely Out-Of-Distribution (OOD) data.
For example, in the case of biosignal classification, models often fail to generalize well on datasets from
different hospitals, due to the distribution discrepancy amongst different sources of data.
Our goal is to demonstrate the Domain Generalization problem present between distinct hospital databases and
propose a method that classifies abnormalities on 12-lead Electrocardiograms (ECGs),
by leveraging information extracted across the architecture of a Deep Neural Network,
and capturing the underlying structure of the signal.
To this end, we adopt a ResNet-18 as the backbone model and extract features from several
intermediate convolutional layers of the network.
To evaluate our method, we adopt publicly available ECG datasets from four sources and handle them as separate
domains.
To simulate the distributional shift present in real-world settings,
we train our model on a subset of the domains and leave-out the remaining ones.
We then evaluate our model both on the data present at training time (intra-distribution) and
the held-out data (out-of-distribution), achieving promising results and surpassing the baseline of a vanilla
Residual Network in most of the cases.
\end{abstract}


\section{Introduction}
During the past decade, there is no doubt that Machine Learning has altered the landscape of industry and
academia.
Since AlexNet's great success story in 2012, deep learning architectures and implementations
have witnessed a Cambrian-like explosion, achieving outstanding results in numerous modalities,
such as Computer Vision, Natural Language Processing, Big Data, etc.
However, despite their success, Deep Neural Networks (DNNs) lack the flexibility of humans, 
as they are highly dependent on the quantity and quality of their training data.
When training such models, we often assume that both the training and evaluation data follow the i.i.d assumption,
i.e., that they are Independent and Identically Distributed.
In real-world settings, however, this assumption often does not hold, as models are applied to data that originate
from different distributions.
As a result, even state-of-the-art methods fail to maintain their accuracy when they are presented
with previously \emph{unseen} data.
In the Machine Learning community, this problem is defined as \emph{Domain Generalization} (DG) and distinct
data distributions are referred to as \emph{Domains}.

\begin{figure}[htbp]
  \centering
  \includegraphics[width=\linewidth]{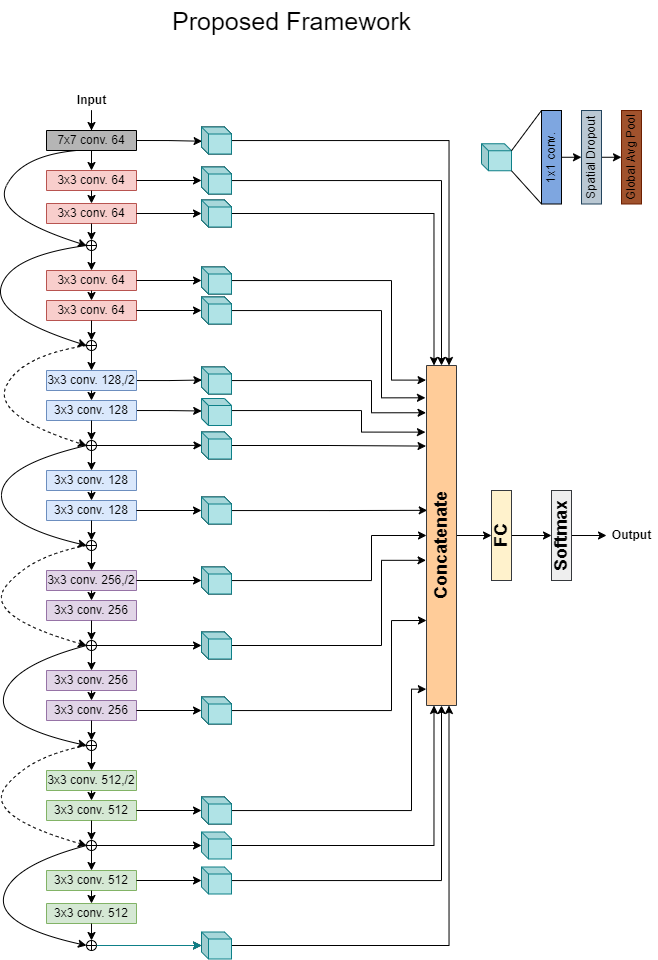}
  \caption{Visualization of the proposed model. The backbone of the architecture
  is a vanilla ResNet-18. By extracting feature maps from intermediate outputs of convolutional layers in the backbone
  model, our method utilizes the CNN's early layers and combines them with features extracted
  from layers further down the network.
  We argue that by incorporating outputs from different intermediate levels of the network, we urge the model
  to disentangle the invariant qualities of the signal.}
  \label{model}
\end{figure}

One field in need of robust models with high generalization abilities, is Medicine.
Medical practitioners are still reluctant in adopting AI-enhanced systems, as methods
which can adequately maintain their performance amongst constantly evolving medical screening methods,
practices and protocols, have yet to be proposed.
Furthermore, the data in most medical databases suffer from \emph{class imbalance}, in the sense that many
diseases are rarely represented in the training data distribution.
Despite the need of generalized models in Medicine, most DG techniques focus on Computer Vision challenges
(e.g object recognition, segmentation or classification).
These methods however, often cannot be applied to other modalities, such as biosignals, due to their dimensionality
and temporal nature.
In this work, we attempt to tackle domain generalization in biomedical
signals, by extracting and combining multiple representations from across the layers of a
Deep Residual Neural Network and classifying 12-lead ECG signals.
We argue that by combining features extracted from multiple layers, the network is able to
infer based on the underlying structure of the signal.
Our contributions can be summarized in the following points:
\begin{itemize}
\item We propose a DG setup for 12-lead multi-label and multi-class ECG classification, using the 6 publicly available datasets in the PhysioNet \cite{alday_classification_2020} database.
\item We demonstrate the inability of a vanilla ResNet-18 to maintain its performance across data domains by evaluating a trained
model on previously unseen ECG data, thus exhibiting the distribution shift present in biosignal datasets.
\item We introduce an alternative architecture which extracts and leverages representations across a Convolutional Neural Network (CNN) and
experimentally demonstrate the improved classification of certain ECG abnormalities for both inter-distribution and out-of-distribution data splits.
\end{itemize}




\subsection{Domain Generalization}
Let $(x, y)$ be a sample and label pair in $(\mathcal{X}, \mathcal{Y})$, drawn
from the (unknown) data distribution.
Domain Generalization algorithms focus on learning a parametric model $f(\cdot; \theta)$,
trained on samples $(x^{(s)},y^{(s)})$ drawn from a set of $N$ \emph{source} Domains $\{S_1, S_2, \dotsc,
S_N\}$, that performs well on data $(x^{(t)}, y^{(t)})$ drawn from $K$
\emph{target} Domains $\{ T_1, T_2, \dotsc, T_K\}$.

In the current work, we consider domain generalization for the 12-lead ECG classification task 
and aim to improve the ability of a model $M_\theta$:
$\mathcal{X}$ $\rightarrow$ $\mathcal{Y}$ to detect domain-invariant features of a class and not be affected
by the distributional shift present in databases from different hospitals.

\section{Previous Work}
In the Machine Learning literature, several previous works have been proposed to mitigate the domain-shift
between training and target domains.
In this section, we briefly summarize the most important past contiributions related to Domain Generalization
in ECG classification.

An abundance of deep learning methods tackling \textbf{ECG classification} have been proposed in the past. From a deep
genetic ensemble of classifiers \cite{10.3389/fncom.2020.564015, plawiak_novel_2020} to LSTMs \cite{gao_effective_2019},
CNNs \cite{9344383, atal_arrhythmia_2020}, and most recently Transformers \cite{9344053}, most methods achieve
promising results when evaluated on data originating from the same distribution of the data they were trained on.
Additional methods such as Self-Supervised Learning \cite{mehari_self-supervised_2021, liu_self-supervised_2021}
and data augmentation \cite{nonaka_randecg_2021} have also been implemented for the extraction
of robust ECG representations.

\textbf{Domain Adaptation} (DA) has made noteworthy progress in the past years.
By taking advantage of pre-trained models on very large datasets, DA algorithms focus on fine-tuning the
network's feature extraction capabilities on previously unseen data or target domains.
The key difference between DA and DG, is that in the case of DA, data from the target distribution are available during
training and are leveraged for the fine-tuning of the model.
With regards to ECG classification, \cite{chen_unsupervised_2020, wang_inter-patient_2021, deng_multi-source_2021}
use \emph{Unsupervised} DA to transfer knowledge acquired  from source domains with a large number of annotated
training examples to target domains with unlabeled data only.

\textbf{Domain Generalization} (DG) methods focus on designing models which can maintain their performance across known and unknown 
data distributions.
In \cite{hasani_classification_2020, shang_deep_2021}, the authors utilize adversarial feature learning
to increase the generalization ability of their models.
To our knowledge, these are the only papers which attempt to describe the DG problem in ECG classification.

\section{Methodology}

Our main hypothesis is that it is challenging to avoid \emph{entangled}
representations, i.e., representations containing both class-invariant information and
domain-specific features when relying only on the last layers of a deep CNN.
We therefore propose to build representations that use features extracted from intermediate layers
of a deep convolutional neural network.

For the backbone model of our methodology we select a ResNet-18 \cite{He_2016_CVPR} and extract features from a total
of 16 layers across the network, as depicted in Fig.1.
To extract said features, we attach a sequential \emph{pipeline} of layers to each of the mentioned 16 layers.
Specifically, the pipeline consists of a 1x1 Convolutional layer, a Spatial Dropout layer and finally a
Global Average Pooling (GAP) layer.
The 1x1 conv layer is used as a projection layer in order to reduce the dimensionality of the extracted feature maps,
leading to a more compressed representation of the signal.
The spatial dropout layer \cite{Tompson_2015_CVPR} of the pipeline, was introduced as a regularizer in an attempt to increase 
the generalization ability of our model, by promoting the independence of each extracted feature map.
Finally, we use a GAP layer to extract the average of each feature map.
The resulting vectors after each pipeline are then concatenated and fed to Fully Connected layer followed by
a Softmax activation.

One of the advantages of the proposed framework is that it is fairly straightforward to implement and the proposed pipelines
can be attached to an arbitrary number of CNN layers.   

\section{Experiments}

\subsection{Datasets \& Preprocessing}
To demonstrate the DG problem in ECG biosignals and to validate our claim that datasets originating from
different hospitals (different data Domains) suffer from distribution shift, we adopt the 6 publicly available
ECG datasets in the PhysioNet database.
Although the datasets are 6 in number, they derive from the following 4 distinct sources and contain a total of
42,585 ECG recordings.
\begin{itemize}
    \item \textbf{CPSC, CPSC Extra}. The first source, is the China Physiological Signal Challenge 2018
    (CPSC2018) \cite{liu_open_2018}.
    \item \textbf{PTB, PTB-XL}. The second source, is the Physikalisch-Technische Bundesanstalt (PTB) Database,
    Brunswick, Germany \cite{wagner_ptb-xl_2020}.
    \item \textbf{INCART}. The third source, is the public dataset from the St. Petersburg Institute of Cardiological Technics (INCART)
    12-lead Arrhythmia Database, St. Petersburg, Russia \cite{tihonenko_st-petersburg_2007}.
    \item \textbf{G12EC}. The fourth source, is the Georgia 12-lead ECG Challenge (G12EC) Database,
    Emory University, Atlanta, Georgia, USA.
\end{itemize}

Since the data originate from different sources, or in the context of DG, from different Domains, the sampling
rate of the signals differ between databases.
As the majority of the signals were sampled at 500Hz and are 10 seconds long, we decided to resample
all recordings at 500Hz and either truncate or zero-pad the remaining ones.
Additionally, since the recordings are multi-labelled more than 100 ECG classes are present across the datasets.
We therefore decided to follow PhysioNet's guidelines and classify only 24 scored classes.
All samples with only unscored labels were removed from the datasets.
We then filter the signals using a 3rd order low-pass Butterworth filter with a cutoff frequency of 20Hz and a
Notch filter with a cutoff frequency of 0.01Hz. After filtering, we normalize the signals between [-1, 1] and
feed them to our model.

\subsection{Experimental Setup}

To demonstrate the DG problem, we consider each data source as a distinct domain and train a vanilla ResNet-18 from scratch
on the datasets from the CPSC and PTB domains, by splitting the data with a 70-10-20 \% train-val-test split ratio.
Inspired by the DG evaluation protocol described in \cite{Li_2017_ICCV}, we evaluate the trained model both on the target
domains INCART and G12EC\footnote{We chose to evaluate on the
INCART and G12EC domains in order to maintain the number of ECG classes 
present in both training and test splits. The class distributions in each dataset can be seen here:
https://github.com/physionetchallenges/evaluation-2020/blob/master/dx\_mapping\_scored.csv. Further information
regarding the datasets can be found at PhysioNet's website: https://moody-challenge.physionet.org/2020/.},
and on the test split of the source domains. 
The same procedure is followed for our implementation.
The source-target domain split can be visualized as in Fig.2.
As previously stated, for the backbone of our model we select a ResNet-18 with randomly initialized weights. 
The selected intermediate layers are depicted in Fig.1 and include all residual half block layers (i.e.
those leading to reduction of the output size) along with selected convolutional layers.
We train our model for 40 epochs and
used Adam as the optimizer, with a batch size of 64 signals.
The initial learning rate is set to 0.001 with a 0.1 decay starting at epoch 24.
We also implement an early stopping protocol, with a patience of 20 epochs.
All models were implemented with Tensorflow and trained on one NVIDIA RTX A5000 GPU\@.

 \begin{figure}[htbp]
\centering
  \includegraphics[width=.7\linewidth]{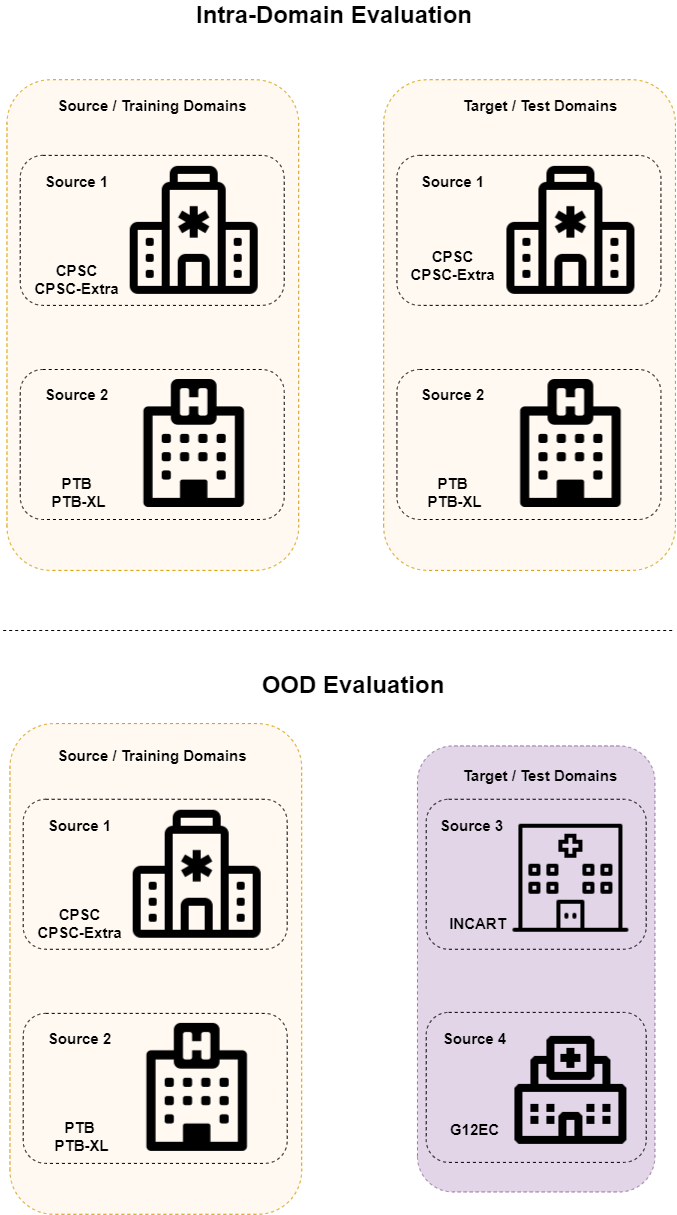}
\caption{To demonstrate the domain shift between datasets from different hospitals, we split the datasets into Source and Target Domains .}
\label{fig:fig}
\end{figure}

\begin{table*}[ht]\centering
\begin{center}
\caption{Intra-distribution and Out-of-Distribution results per ecg class. We compare our implementation
with the baseline model and evaluate on intra and OOD data distributions. We choose to omit the classes which were not
classified by either the baseline or our implementation. The best intra-distribution results
are highlighted in $\textbf{bold} $ and the best OOD results are \underline{underlined}.
For simplicity, we only compare the per class F1-scores.}
\label{tab:metrics}
\begin{adjustbox}{width=1\textwidth}
\begin{tabular}{|c||ccc|ccc||ccc|ccc|}
\hline
\multicolumn{13}{|c|}{\hspace{5cm} Intra-Distribution \hspace{7.3cm} OOD} \\
\hline
&  & Baseline &  &  & Our Model &  &  & Baseline &  &  & Our Model &\\
Diagnosis & Precision & Recall & F1-Score & Precision & Recall & F1-Score & Precision & Recall & F1-Score & Precision & Recall & F1-Score\\

\hline

atrial fibrillation & 0.80 & 1.00 & 0.89 & 0.89 & 1.00 & \textbf{0.94} & 0.80 & 0.80 & 0.80 & 0.83 & 1.00 & \underline{0.91} \\
t wave abnormal & 0.50 & 0.50 & 0.50 & 0.75 & 0.75 & \textbf{0.75} & 0 & 0 & 0 & 0 & 0 & 0 \\
1st degree av block & 0.50 & 0.60 & 0.54  & 0.57 & 0.8 & \textbf{0.67} & 0.50 & 0.40 & 0.44 & 0.67 & 0.80 & \underline{0.73} \\
left axis deviation & 0.75 & 0.23 & 0.35   & 0.73 & 0.85 & \textbf{0.79} & 0 & 0 & 0 & 0.60 & 0.30 & \underline{0.40} \\
sinus bradycardia & 0 & 0 & 0 & 0 & 0 & 0  & 0 & 0 & 0 & 1.00 & 0.10 & \underline{0.20} \\
sinus rhythm & 0.84 & 0.97 & 0.90   & 0.92 & 0.95 & \textbf{0.94} & 0.37 & 0.93 & 0.53 & 0.48 & 0.80 & \underline{0.60} \\
sinus tachycardia & 1.00 & 0.75 & 0.86   & 1.00 & 1.00 & \textbf{1.00} & 0.89 & 0.89 & \underline{0.89} & 0.87 & 0.78 & 0.82 \\
left anterior fascicular block & 1.00 & 0.6 & \textbf{0.75}   & 0.60 & 0.60 & 0.60 & 0 & 0 & 0 & 0 & 0 & 0 \\
supraventricular premature beats & 0 & 0 & 0   & 0.17 & 0.33 & \textbf{0.22} & 0.33 & 0.20 & 0.25 & 0.33 & 0.40 & \underline{0.36} \\
nonspecific intraventricular conduction disorder & 0 & 0 & 0  & 1.00 & 0.25 & \textbf{0.40} & 0 & 0 & 0 & 0 & 0 & 0 \\
incomplete right bundle branch block & 1.00 & 0.50 & \textbf{0.67}  & 1.00 & 0.25 & 0.40 & 0 & 0 & 0 & 0 & 0 & 0\\
complete right bundle branch block & 0.75 & 0.50 & \textbf{0.60}  & 1.00 & 0.33 & 0.50 & 0.41 & 0.87 & 0.56 & 0.41 & 0.87 & 0.56 \\

\hline
\end{tabular}
\end{adjustbox}
\end{center}
\end{table*}

\subsection{Results}

To properly demonstrate the consequences of the domain shift, we choose to evaluate our implementations
on each class separately.
As the datasets suffer from class imbalance, we choose to adopt the per class \emph{Precision, Recall and F1-score} evaluation metrics, as 
\emph{Accuracy} can often be misleading.
Additionally, we omit the classes which were not recognized by neither the baseline model or
our framework.
Out of 24 labels, only 12 were classified by either the baseline or our method.
Table 1 depicts the experimental results for both models.

As depicted in Table 1, when comparing the OOD evaluations of both implementations with their intra-distribution
performances, the per-class drop in the F1-score is transparent, thus demonstrating the presence of the
DG problem in ECG datasets.
However, even though our model is in some cases clearly affected by the domain-shift between source and target data,
it performs close to  or even exceeds its intra-distribution performance, in the cases of atrial fibrillation
1st degree av block and supraventricular premature beats respectively.

Compared to the vanilla ResNet-18 model in the intra-distribution evaluation, our method is able to surpass the
baseline in 6 out of 10 commonly predicted classes and is even capable of recognizing 2 extra classes.
A similar trend follows when comparing the OOD evaluations, as our model surpasses the baseline in 4 out of 6 common
classes, performs the same on 1 common class and recognizes 2 extra classes.

The experimental results seem to support our claim, that our model exhibits generalization capabilities when presented
with OOD data, as it is able to maintain its performance in several cases and even infer classes which were not
perceived by its baseline counterpart.
Moreover, our framework also surpasses the baseline in the classic intra-distribution evaluation, in the
majority of the cases.

\section*{Conclusion}

In this paper we attempt to tackle the \textit{Domain Generalization} problem in ECG classification by combining
multiple representations from across the architecture of a Deep Convolutional Neural Network.
To validate our claim, we initially demonstrate the distributional shift present in ECG databases which originate
from distinct sources, by experimentally displaying the performance drop of trained models when assessed on OOD data.
We argue that by utilizing extracted features from a CNN’s intermediate layers, the model can be forced
to incorporate invariant features in the biosignal representation.
When evaluating on intra-distribution and OOD data, our model indeed displays generalization capabilities as it maintains
its performance across domains and surpasses the baseline model, in several cases.
As future work, we aim to build upon our initial method and attempt to explicitly constrain the model into extracting
the invariant features of one-dimensional biosignals, by adding additional regularization terms.
Futhermore, we also intend to dive into our model's results by visualizing its inference mechanisms
with saliency maps.

\bibliographystyle{IEEEtran}
\bibliography{Library}
\vspace{12pt}
\end{document}